\documentclass[runningheads]{llncs}
\usepackage[T1]{fontenc}
\usepackage{cite}
\usepackage{graphicx}
\usepackage{amsmath,amssymb,amsfonts}
\usepackage{xcolor}
\usepackage[caption=false,font=normalsize,labelfont=sf,textfont=sf]{subfig}
\usepackage{times}
\usepackage{algorithm}
\usepackage{algorithmicx}
\usepackage{algpseudocode}
\floatname{algorithm}{Algorithm} 
\usepackage{multirow}
\usepackage{booktabs}
\newtheorem{assumption}{Assumption}
\usepackage{multicol}
\usepackage{subcaption}
\usepackage{marvosym}

\begin{document}
\title{Personalized federated prototype learning in mixed heterogeneous data scenarios}
%

\author{Jiahao Zeng\inst{1,2}\and Wolong Xing\inst{1,2}
\and Liangtao Shi \inst{3}\and Xin Huang \inst{1,2}\and Jialin Wang \inst{1,2}\and \\
Zhile Cao\inst{1,2}\and Zhenkui Shi\inst{1,2}\textsuperscript{(\Letter)}}
\authorrunning{Zeng. Author et al.}
%
\institute{ Guangxi Normal University, Guilin, 541004, China \\
\and Key Lab of Education Blockchain and Intelligent Technology, Ministry of Education, GuangXi Normal University, Guilin, 541004, China\\
\and
Hefei University of Technology, Hefei, 230009, China\\
\email{shizhenkui@gxnu.edu.cn}}
\maketitle              
\begin{abstract}

Federated learning has received significant attention for its ability to simultaneously protect customer privacy and leverage distributed data from multiple devices for model training. However, conventional approaches often focus on isolated heterogeneous scenarios, resulting in skewed feature distributions or label distributions. Meanwhile, data heterogeneity is actually a key factor in improving model performance. To address this issue, we propose a new approach called PFPL in mixed heterogeneous scenarios. The method provides richer domain knowledge and unbiased convergence targets by constructing personalized, unbiased prototypes for each client. Moreover, in the local update phase, we introduce consistent regularization to align local instances with their personalized prototypes, which significantly improves the convergence of the loss function. Experimental results on Digits and Office Caltech datasets validate the effectiveness of our approach and successfully reduce the communication cost.

\end{abstract}
\section{Introduction}

The rapid proliferation of mobile phones, wearables, tablets, and smart home devices has led to exponential growth in the volume of data generated and retained by these devices. These data contain valuable insights for device owners. However, many users have become increasingly concerned about privacy, demanding that their data remain exclusively on local devices. Federated Learning (FL) \cite{10.1145/3298981} provides a privacy-preserving distributed machine learning framework. In FL, a cloud server coordinates with distributed clients while ensuring data privacy through localized storage. The foundational FedAvg algorithm \cite{mcmahan2017communication} iteratively aggregates client model parameters and distributes averaged global models to clients, enabling collaborative training without privacy disclosure. However, real-world scenarios involve data from heterogeneous sources with distinct characteristics, resulting in non-independent and identically distributed (non-IID) data \cite{li2020federated,tan2023federated}. Local client updates based on their data distributions often diverge from the global optimization trajectory. Personalized Federated Learning (PFL) has emerged as a prominent approach to developing client-specific models tailored to individual data distributions.

Personalized Federated Learning (PFL)~\cite{xing2024personalized} addresses data heterogeneity by enabling clients (e.g., mobile devices or organizations) to develop customized models aligned with their unique data distributions \cite{10689471}. Two fundamental challenges persist: (1) label distribution skew across clients, and (2) feature distribution divergence within identical label classes. Existing research predominantly focuses on single-mode heterogeneity (either label or feature skew), with limited exploration of cross-domain mixed heterogeneity where data originates from divergent domains with varying label distributions.

Under label skew conditions, the global model exhibits bias toward majority classes, leading to suboptimal generalization of personalized models on local client data. While hybrid local-global optimization \cite{hanzely2020federated} and model decoupling techniques \cite{li2020prototypical} show efficacy in handling label skew, they fail to address feature distribution bias as the global model struggles to capture client-specific feature representations—even for data instances sharing identical labels across clients \cite{10795221}. This feature space misalignment further hinders effective inter-client model collaboration. In practical cross-domain deployments with dual heterogeneity (concurrent label skew and feature divergence), these limitations not only degrade model performance but also hinder real-world applicability. Consequently, developing unified solutions for hybrid heterogeneous scenarios becomes imperative.

Real-world applications frequently exhibit dual heterogeneity scenarios combining label distribution skew and feature distribution divergence, as exemplified by cross-institutional CT image analysis where hospitals in different geographic regions possess distinct patient cohorts. This dual heterogeneity arises from two primary factors: (1) feature variations caused by discrepancies in medical imaging equipment specifications, and (2) label distribution skew stemming from demographic differences in disease prevalence across hospital populations.

Building upon prototype learning foundations \cite{li2020prototypical,zhu2021prototype}, we present Prototypical Federated Partial Learning (PFPL), a novel framework for hybrid heterogeneity scenarios. PFPL employs cross-domain Lipschitz-constrained prototype comparison to quantify domain-specific knowledge relevance. Through adaptive prototype aggregation weighted by inter-domain similarity metrics, it constructs client-specific prototypes that mitigate dominant domain bias. Furthermore, we introduce Personalized Prototype Alignment (PPA), a regularization mechanism that enforces consistency between local instance embeddings and client-specific prototypes through feature-space distance minimization, ensuring robustness under hybrid heterogeneity. The main contributions of this paper are summarized as follows:
\begin{itemize}
    \item[$\bullet$] We propose a novel personalized prototype learning approach aimed at solving the problem of skewed label distribution and skewed feature distribution in hybrid heterogeneous scenarios.
    \item[$\bullet$] To cope with the label distribution imbalance problem, we introduce prototype learning to capture domain knowledge and propose a novel aggregation scheme to generate personalized prototypes for each client. Meanwhile, in the local update phase, we design personalized unbiased prototype consistency to provide fair and unbiased target signals by narrowing the feature distance between instance embeddings and personalized prototypes, thus effectively mitigating the impact of feature distribution imbalance on model performance.
    \item[$\bullet$] We conduct extensive experiments on Digits and PACS tasks. The experimental results show that our scheme outperforms some recent federated learning methods in call and heterogeneous scenarios.
\end{itemize}

\begin{figure}
\centering 
\includegraphics[width=\textwidth]{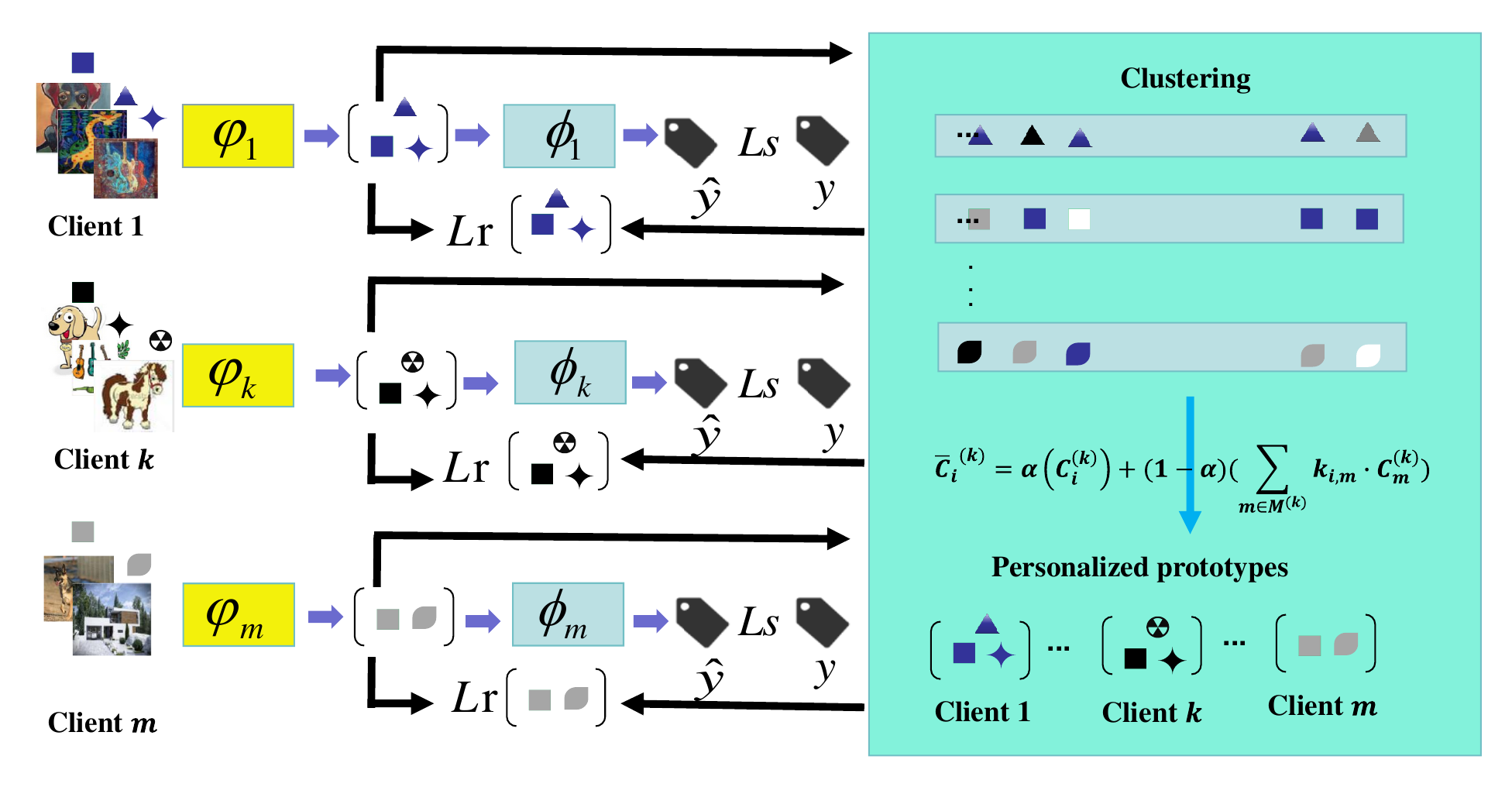} 
\caption{Architecture of personalized Federated Prototype Learning (PFPL). 
} 
\label{fig_1}
\end{figure} 

\section{Related work}
\label{sec:2}
\subsection{Heterogeneous challenges in federated learning}

Data heterogeneity in federated learning primarily manifests as two distinct types \cite{li2021fedbn,huang2023rethinking}: label distribution skew and feature distribution shift \cite{YI2024111633}. To address label distribution skew, conventional methods often employ label-based dataset partitioning to construct pseudo-IID distributions, aiming to reduce training bias and enhance model generalization. pFedKT \cite{YI2024111633} achieves personalized-generalized balance through dual knowledge transfer: (1) local hypernetworks preserve historical personalized knowledge, and (2) contrastive learning propagates updated global knowledge. Other methods like FedProto \cite{dai2023tackling} and FedProc \cite{mu2023fedproc} enforce feature-level consistency through prototype alignment and procedural feature matching, respectively. However, these methods predominantly focus on single-domain label skew scenarios while neglecting cross-domain feature shifts in real-world hybrid heterogeneity.

Feature distribution shift poses a distinct challenge, where cross-domain client data leads to suboptimal cross-domain generalization \cite{Chen_2023_ICCV}. FedBN \cite{li2021fedbn} addresses feature shift through client-specific batch normalization layers prior to model aggregation. ADCOL \cite{li2023adversarial} and FCCL \cite{huang2022learn} impose substantial resource overhead, requiring adversarial discriminators and public datasets for cross-client alignment. While FPL \cite{huang2023rethinking} mitigates feature shift via prototype clustering, it prioritizes global model convergence over client personalization. These approaches primarily target isolated feature shift scenarios while neglecting concurrent label distribution skew.Our work addresses hybrid heterogeneity—simultaneous label skew and feature shift—by developing personalized models tailored to individual client data characteristics.


\subsection{Personalized federated learning}
Personalized federated learning is extensively employed to address the data heterogeneity issue in federated learning. This approach enables each client to customize and optimize the personalized model in accordance with the characteristics and requirements of its local data, thereby facilitating more precise localized model training and adaptation.

Personalized federated learning has evolved diverse architectural strategies to address data heterogeneity. Among parameter decoupling approaches, FedRep \cite{husnoo2022fedrep} implements a two-phase local update protocol: (1) clients first train personalized classification heads with frozen feature extractors (bodies) from the server, followed by (2) body parameter refinement using updated head parameters. Filip Hanzely et al. \cite{li2020prototypical} have proposed a method that generates personalized models for each client by mixing local and global models to balance the two. FedBABU \cite{oh2021fedbabu} extends this paradigm through a three-stage process: local body training with fixed random-initialized heads, server-side body aggregation, and post-training head fine-tuning for personalization. FedRoD \cite{li2023adversarial} innovates further with a dual-head architecture comprising a shared general head optimized via class-balanced loss and client-specific private heads trained with empirical loss, where only the body and general head participate in aggregation. Prototype-enhanced methods offer complementary solutions. FedNH \cite{dai2023tackling} integrates prototype-semantic consistency learning to enhance feature discriminability while employing head regularization to prevent prototype collapse under class imbalance. This framework adopts alternating optimization: frozen-body head updates precede fixed-head body refinements.

However, most of the above personalization methods only consider the heterogeneous problem for a single scenario (skewed label distribution or skewed feature distribution). There are relatively few personalization methods for mixed scenarios of both, which limits the application of federated learning on more diverse non-IID data. Therefore, solving more diverse hybrid heterogeneous problems has become an important challenge for federated learning research.


\begin{figure}
\centering 
\includegraphics[width=\textwidth]{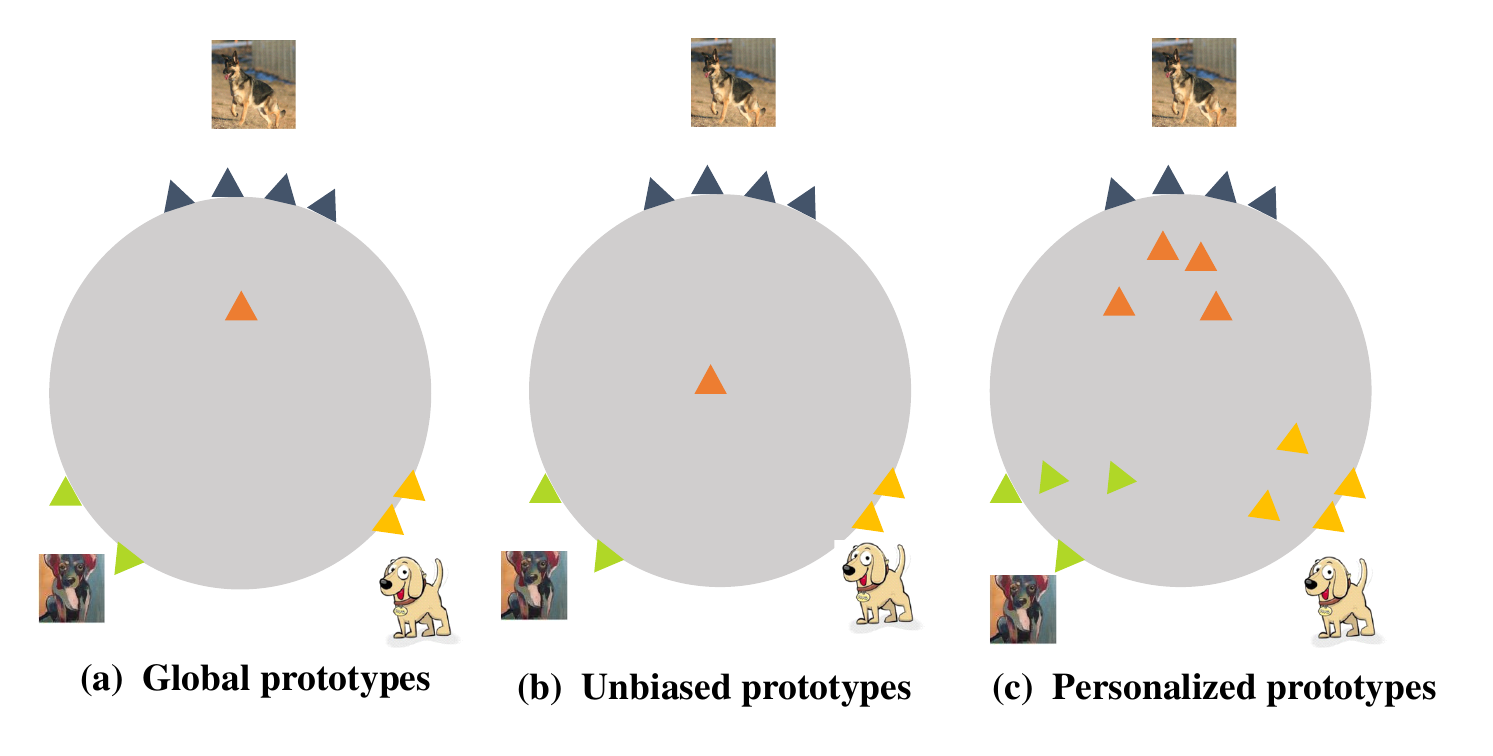} 
\caption{Description of different prototypes.}
\label{fig_2}
\end{figure} 

 \section{Methodology}
\label{sec:3}
\subsection{PRELIMINARY}
Following typical federated learning \cite{10.1145/3298981}, there are $M$ participants, and the private data set of the participants is $D_m=\{x_i,y_i\}_{i=1}^{N_m}$, where $N_m$ represents the client side data size. These private data follow different label distributions and come from different domains. For example, data sets $D_i$ and $D_j$ on two client sides $i$ and $j$ may have different label statistical distributions. This is common for photo classification apps installed on the mobile client side. The server needs to identify many classes $\mathbb{C}=\{\mathbb{C}(1),\mathbb{C}(2),\cdots \mathbb{C}(k),\cdots\}$, while each client side only needs to identify a few classes that make up a subset of $\mathbb{C}$. The class sets may vary from client side to client side, although there is overlap. And these private data sets are derived from different domains, resulting in significant differences in the features of the data even if the categories are the same.

\begin{itemize}
    \item[$\bullet$] \textbf{Mixed heterogeneous scenarios in federated learning:} $ P_i (x \vert y)\neq P_j(x \vert y),(P_i (y)\\
    \neq P_n (y)).$ There is a skewed feature distribution and a skewed label distribution between private data. Specifically, the label distribution before different client sides is different, and the data comes from different domains, presenting a unique feature distribution despite the overlap between domains.
\end{itemize}

In addition, participants agree to share models with the same architecture. We treat the model as two modules:
 Feature Extraction Module $\phi$ (i.e., the embeddings function) transforms the input from the raw feature space to the embedding space, $f(\phi,x) \to h\in R^d$, the sample$ x$ coding d-dimensional feature vector $ h = f (\phi,x)\in R^d$. Decision Module $\varphi$ makes classification decisions for the given learning task. $g:(\varphi,h)\to \hat{y} \in \mathbb{C} $ maps feature $ h$ to the logits output $\hat{y} = g (\varphi,h)\in \mathbb{C}$. So, the label function can be written as: $F(\phi,\varphi)=f(\phi)\cdot g(\varphi)$, and we use $\omega$ to represent $(\phi, \varphi)$ for short.

 \textbf{Prototypes:} Each prototype is the average of the feature vectors of the same class:
\begin{equation}
\label{eq:1}
     C^{(k)}=\frac{1}{|{ D^k} |} \sum_{(x,y )\in D^k}f(\phi;x).
\end{equation}

 where $D^{k}$ represents the  data instances label $ K$, and $|{ D^{k}} |$ represents the number of data instances label $K$.

 \textbf{Local Prototypes:} We define a feature $C^{(k)}$ to represent the $k$-th class in $\mathbb{C}$. For the $i$-th client, $C^{(k)}$ is the average of the features obtained by inputting samples with the label $k$ into the feature extraction module.
\begin{equation}
\label{eq:2}
    C_i^{(k)}=\frac{1}{|{D_{i}^{k}} |} \sum_{(x,y )\in D_{i}^{k}}f_{i}(\phi_{i};x).
\end{equation}

where $D_{i}^{k}$ represents the data samples of class $K$ in the $i$-th client.

\textbf{Global Prototypes:} For a given class $j$, the server receives locally computed features with class label $j$ from a group of clients. These local features with labels $j$ are aggregated by taking their average to generate the global feature $\overline{C}^{(j)}$ for class $j$.

\begin{equation}
\label{eq:3}
    \overline C ^{(k)} = \frac{1}{\aleph^k} \sum\limits_{i \in {\aleph^k}} {\frac{|D_{i}^{k}|}{N^k}} C_{i}^{(k)}.
\end{equation}

where $C_i^{(k)}$ represents the local features of class $K$ from the $i$-th client, $\aleph^k$ represents the set of clients that have class $K$, and $N^k $represents the total number of data instances of all client-side classes label $K$.

However, the global prototype is not suitable for the mixed heterogeneous scenario in this paper, which mainly has two problems: \textbf{1}. A single global prototype blurs the difference between different domains, and it is difficult to learn special knowledge between different domains. \textbf{2}. Since the weight parameter of the global prototype is determined by the amount of data in the category sample, the final global prototype is biased towards the dominant user with a large amount of data, which makes it difficult for the client side with few data instances to learn. A simple approach is to build an unbiased prototype; that is, the prototype weight of each client is the same, but this approach still faces the challenge of problem 1, and this approach makes the client side with less sample instances benefit but hurts the client side with more sample instances to participate in federated learning.

\subsection{Personalized federated prototyping learning}

We propose a solution for hybrid heterogeneous FL. This paper uses a prototype as the main component for exchanging information at the client side and server level. The framework is shown in Figure~\ref{fig_1}. The central server receives local prototype sets $\{ \Theta(1), \Theta(2), \cdots, \Theta(m)\}$ from $ m$ local client sides and then clusters prototypes $\{\Phi(1),\\\Phi(2),\cdots, \Phi(k),\cdots\}$ of the same category. In a hybrid heterogeneous FL setup, these prototype sets overlap but are not the same. Take the handwritten digit data set as an example. The first client side is the recognition numbers 2, 3, 4, from the MNIST data set, while the other client side is the recognition numbers 4, 5, from the SYN data set. These are two sets of handwritten digits from different domains, with different sample categories, albeit overlapping. For the prototype category of each client in the cluster, by assigning weights with the L2 distance of the other client-side prototypes, aggregation generates a personalized prototype specific to the client.

\textbf{Personalized prototype:} The server level collects the prototype set of the client and clusters it for the client side prototype category. Taking the prototype with the category $K$ of client side $i$ as an example, its personalized prototype is as follows:

\begin{equation}
\label{eq:4}
    \overline {C}_{i}^{(k)}=\alpha(C_{i}^{(k)})+(1-\alpha)(\sum_{m\in M^{ (k)}}k_{i,m}\cdot C_{m}^{(k)} )
\end{equation}

where $\alpha$ represents a hyperparameter that controls the degree of personalization, $M^{(k)}$ represents a client side cluster with a $ K$-class label prototype, $C_{m}^{k}$ represents the $K$-class label prototype uploaded by the $m$ client side, and $k_{i,m}$ represents the weight coefficient of the client side $m$ to the $i$ client. The calculation formula is determined by comparing the L2 distance of the two client side $K$-class prototypes, as follows  $k_{i,m}=\frac{\Vert C_{i}^{(k)},C_{m}^{(k)}\Vert_2}{\sum_{m\in\Phi(k)}\Vert C_{i}^{(k)},C_{m }^{(k)}\Vert_2}$, where $\Phi(k)$ represents the cluster prototype with the server level label $K$, and the L2 distance between prototypes is calculated as $\Vert C_{i}^{(k)},C_{m}^{(k)}\Vert_2=\sum_{k} d(C_{i}^{(k)}-C_{m}^{(k)})^2$, Where $d$ is the locally generated distance metrics of prototype $C_{i}^{(k)}$ with label $k$ and prototype $C_{m}^{(k)}$ with the same label on the other client side $m$. Distance measures can take many forms, such as L1 distance, L2 distance, and bulldozer distance. Here we use L2 distance metrics.

We generate its personalized prototype for each client side. In simple terms, personalized prototypes with the same label on different clients are affected by domain migration differently. When assigning weight, prototypes from the same domain will assign more weight, while the weight assigned from different domains will be less, making personalized prototypes tend to be more knowledge of the same domain and stay away from the influence of different domains, thus effectively solving the above problem \textbf{1} and our weight is determined according to the L2 distance between different client side prototypes, and is not determined by the amount of data on the client side, so it will not be affected by the dominant domain. Although the amount of data on individual prototypes is small, it can also learn more from a large number of prototypes of the same domain, thus solving the problem \textbf{2}.

\textbf{Local Model Update:} The client side needs to update the local model to generate consistent client-side functionality. We also introduce unbiased personalized prototype consistency, which allows the local prototype to approximate its personalized prototype through regularization in local updates. Specifically, the loss function is defined as follows:

\begin{equation}
\label{eq:5}
    \ell (D_i,w_i)=\ell _S(F(w_{i};x_i),y_i) +\\ \lambda \cdot  \ell_R\left
    ( \overline {C}_{i}^{(k)},C_i^{(k)} \right )
\end{equation}

where $D_i$ stands for data from the $i$-th client, $\lambda$ is an important parameter for regularization.

\begin{algorithm}[H]

        \caption{PFPL} 
        \begin{algorithmic}[1] 
            \Require $D_i$, $w_i$, $i=1,\cdots m$
            \Ensure The final personalization model $\{w_i\}$, $i=1,\cdots,m$
            
            \State \textbf{Server executes: ($\Theta(i),\overline { C } _ { i } ^ { (k ) }$)}
            \State \textbf{Initialize} $w$ for all clines
            \For{each round $T=1,2,\cdots$} 
                \For{each client $i$ in parallel }     
                    \State $\{\Theta(i)\}\longleftarrow$ LocalUpdate $        ( i,\overline C ^{(k)} )$ 
                    
                \EndFor
    
                \State Clustering prototype sets  uploaded by clines
                \State Update personalized prototype sets by :
                \State $\overline { C } _ { i } ^ { (k ) } = \alpha ( C _ { i } ^ { k} ) + ( 1 - \alpha ) ( \sum _ { m \in M ^ { ( k ) } } k _ { i , m } \cdot C _ { m } ^ {k } )$
                \State Send the personalized prototype to the corresponding
                \State client side
                
            \EndFor
            \State\textbf{LocalUpdate ($i,\overline { C } _ { i } ^ { (k ) } $):}
            \For{each local epoch }
                \For{batch  $(x_i,y_i)\in D_i$}
                \State Compute local features by Eq.~\ref{eq:2}
                 \State Compute loss by Eq.~\ref{eq:5} using local prototype.
                  \State Update local model according to the loss. 
                  \State Update local prototype sets $\Theta(i)$ with personalized 
                \State prototypes in $\{\overline C_i ^{(k)}\}$
                \EndFor{}
            
            \EndFor{}
            \State \Return $\Theta(i), i=1,...m$
        \end{algorithmic}
        \label{algorithmic1}
        
\end{algorithm}

\textbf{Discussion:} We further explain the differences between the three prototypes in Figure ~\ref{fig_2}. The global prototype inherently confuses the knowledge of different domains and shows a skewed feature space towards the potentially dominant domain in heterogeneous federated learning. Unbiased prototypes also have the problem of confusing the knowledge of different domains, and giving the same weight to the client side with a large number of instance samples is itself an unfair allocation, which reduces the enthusiasm of the client side with a large data instance to participate in federated learning. Our personalized prototype solves the above two problems at the same time. Specifically, for problem 1, our personalized prototype is generated for different client side aggregates rather than a single global prototype or unbiased prototype, which effectively solves the problem that the client side data comes from different domains. For problem 2, our weight coefficient is mainly determined by hyperparameter a and K, a guarantees the degree to which our personalized prototype is biased towards the local prototype. The calculation formula of K guarantees that other client-side prototypes from the same domain assign more weight, while client-side prototypes from different domains assign less weight, so that the final personalized prototype is biased towards the real domain and deviates from other domains. Compared with the traditional model gradient parameter, the dimension of the prototype is much smaller than that of the overall model, which brings less computational cost to the participants. In addition, prototype uploads are privacy-safe because they are one-dimensional vectors generated by averaging low-dimensional representations from the same class of samples, which is an irreversible process. Second, an attacker cannot rebuild the original data source from the prototype without accessing the local model. Therefore, prototype not only provides lower computational cost, but also a privacy-preserving scheme in heterogeneous federated learning.

\subsection{Optimization Objective}

The goal of PFPL is to solve joint optimization problems on distributed networks. PFPL applies prototype-based communication, which allows a local model to align its local prototype with its personalized prototype while minimizing the sum of losses for all client side local learning tasks. The learning goal of personalized federated prototypes across heterogeneous clients can be expressed as:

\begin{equation}
\begin{aligned}
\mathop {\arg \min }\limits_{\phi, \varphi }  \sum\limits_{i = 1}^m \frac{{{D_i}}}{N}\ell _S(f(w_{i};x_i),y_i) +\\\lambda\cdot \sum\limits_{k= 1}^{|\mathbb{C} |} \sum\limits_{i = 1}^m  \ell_R\left ( \overline{C}_i^{(k)} ,C_i^{(k)} \right ).
\end{aligned}
\label{eq:6}
\end{equation}

where loss $\ell _S(F(w{i};x_i), y_i)$ represents the objective loss for the $i$-th client, and we use the standard cross-entropy loss as the objective loss function. $N$ represents the sum of all client-side instance data, $\vert\mathbb{C}\vert$ represents the number of classes for the labels, and $\ell_R$ is the regularization term used to measure distance, with its expression as follows:

\begin{equation}
    \ell _R\left (\overline{C}_i^{(k)} ,C_i^{(k)} \right )=\Vert  \overline{C}_i^{(k)},C _ { i } ^ {(k )}\Vert_2
\label{eq:7}
\end{equation}

where $\ell _R$ is the distance metrics of the locally generated prototype $C_i^{(k)}$ and the globally aggregated personalized prototype $\overline{C}_i^{(k)}$. Here we use the L2 distance to measure the difference between the two.

\section{Convergence analysis}
\label{sec:4}
We use the first-level model (decision module) as our objective loss function.

\begin{assumption}
(Lipschitz Smooth). It is assumed that each local objective function is $L_1$-Lipschitz Smooth, which also implies that the gradient of the local objective function is $L_1$-Lipschitz continuous.
\label{assumption1}
\end{assumption} 

\begin{equation}
    \begin{aligned}
         \left \|  \bigtriangledown \ell_ {t_{2}} -\bigtriangledown\ell_{t_{2}}  \right \|_{2}\le
 L_1 \left \| w_{i,t_{1}}-w_{i,t_{2}} \right \|_{2},\\ \forall t_1,t_2>0,i\in \{1,2,\cdots, m\}.
    \end{aligned}
\label{eq:8}
\end{equation}

This also implies the following quadratic bound:

\begin{equation}
    \begin{aligned}
        \ell_ {t_{1 }} -\ell_{t_{2}}  \le  \left \langle \bigtriangledown \ell_ {t_{2}},(w_{i,t_{1}}-w_{i,t_{2}}) \right \rangle +\\ \frac{L_1}{2} \left \| w_{i,t_{1}}-w_{i,t_{2}} \right \|_{2}^2,
        \forall t_1,t_2>0,i\in \{1,2, \cdots, m\}.
    \end{aligned}
\label{eq:9}
\end{equation}
  
\begin{assumption}
(Unbiased Gradient and Bounded Variance)
The stochastic gradient $g_{i,t}=\ell(w_{i,t})$ is an unbiased estimator of the local gradient for each client. Assuming that its expectation satisfies the following equation:
\label{assumption2}
\end{assumption} 

\begin{equation}
    \begin{aligned}
        E_{{\xi}_i}\sim _{D_i}\left [ g_{i,t} \right ] =\bigtriangledown \ell(w_{i,t})=\bigtriangledown \ell_t,\\\forall i\in \{1,2, \cdots, m\},
    \end{aligned}
\label{eq:10}
\end{equation}

and its variance is bounded by  $ \sigma ^2$:
\begin{equation}
    E \left [ \left \| g_{i,t}-\bigtriangledown \ell_{(w_{i,t})} \right \| _2^2 \right ]  \le \sigma ^2.
\label{eq:11}
\end{equation}

\begin{assumption}\label{assumption3}
(Bounded Expectation of Euclidean norm of Stochastic Gradients). The expectation of the random gradient is bounded by $G$:
\end{assumption} 

\begin{equation}
    E \left [ \left \| g_{i,t} \right \| _2 \right ]  \le G,\forall i\in \{1,2, \cdots, m\}.
\label{eq:12}
\end{equation}

\begin{assumption}\label{assumption4} The functions of each feature extraction module, commonly known as embedding functions, are $L_2$-Lipschitz continuous.
\end{assumption}

\begin{equation}
    \begin{aligned}
        \left \| f_i(\phi _{i,t_1})- f_i(\phi _{i,t_2})  \right \|\le 
L_2\left \|\phi _{i,t_1}-\phi _{i,t_2}  \right \| _2,\\\forall t_1,t_2>0, i\in \{1,2, \cdots, m\}.
    \end{aligned}
\label{eq:13}
\end{equation}
   
We can obtain theoretical results for non-convex problems if the above assumption holds. In Theorem \ref{theorem1}, we provide the expected decrease in each round. We use $e \in \{\frac{1}{2}, 1, 2, \cdots, E\}$ to represent local iterations and $t$ to represent global communication rounds. Here, $tE$ represents the time step before global features aggregation, and $tE +\frac{1}{2}$ represents the time step between global features aggregation and the first iteration of this round.

\begin{theorem}\label{theorem1}
(One-round deviation) Let Assumption \ref{assumption1} to \ref{assumption4} hold. For an arbitrary client, after every communication round, we have,
\end{theorem} 

\begin{equation}
    \begin{aligned}
        \left [ \left \| \ell _{(t+1)E+\frac{1}{2}} \right \|  \right ] \le \ell _{tE+\frac{1}{2}}-\left ( \eta -\frac{L_{1} \eta^2}{2}  \right ) \\\sum_{e=\frac{1}{2}}^{E-1} \left \| \bigtriangledown \ell _{tE+e} \right \| _2^2+\frac{L_{1}E \eta^2}{2}\sigma ^2+\lambda L_2 \eta EG.
    \end{aligned}
\label{eq:14}
\end{equation}

Theorem \ref{theorem1} indicates the deviation bound of the local objective function for an arbitrary client after each communication round. Convergence can be guaranteed when there is a
certain expected one-round decrease, which can be achieved
by choosing appropriate $\eta$ and $\lambda$ .

\begin{corollary}\label{corollary1}
(Non-convex pFedPM convergence). The
loss function $\ell$ of an arbitrary client monotonously decreases in every communication round when
\end{corollary} 

\begin{equation}
    \begin{aligned}
        \eta_{\acute{e}}  < \frac{2( {\textstyle \sum_{e=\frac{1}{2}}^{\acute{e}}} \left \|\bigtriangledown \ell _{tE+e}\right \| _2^2-\lambda L_2EG ) }{L_2  EG} ,
    \end{aligned}
\label{eq:15}
\end{equation}

where  $\acute{e} = \{\frac{1}{2}, 1, 2,\cdots , E \}$.and 

\begin{equation}
    \lambda_t < \frac{\left \|\bigtriangledown \ell _{tE+\frac{1}{2}}\right \| _2^2  }{L_2  EG}.
\label{eq:16}
\end{equation}

Thus, the loss function converges.

Corollary \ref{corollary1} guarantees that the expected bias of the loss function is negative, ensuring the convergence of the loss function. We can further ensure the convergence of the algorithm by choosing appropriate learning rates $\eta$ and importance weights $\lambda$.

\begin{table*}[t]
    \caption{\centering Results for the Digits dataset on different algorithms. 
    }
    \centering
     \resizebox{\linewidth}{!}{
    \begin{tabular}{c|ccccc|c|c|cc}
     \midrule
     \multirow{2}{*}{Algorithms} &
      \multicolumn{5}{c|}{Acc n=3} & Acc n=4 & Acc n=5 &  
      \multirow{2}{*}{Rounds} 
      \\
      \quad  & MNIST & USPS & SVHN & SYN & AVG & AVG & AVG & \quad  \\
 \midrule
        Local  &  98.32 & 93.64 & 85.42 & 53.18 & 91.15 &  92.17 &  92.86 &  0  \\ 
        FedAvg  & 98.64 & 92.17 & 86.35 & 54.16 & 87.56 &  88.42 & 88.75 & 200 \\ 

        Ditto  &  96.35 & 90.86 & 85.44 & 53.12 & 86.74 & 87.14 &  87.56 & 200  \\ 
        APFL  &  98.42 & 91,24 & 86.14 & 53.42 & 87.98 &  88.58 &  89.43 & 200  \\ 
        FedRod  & 95.58 & 90.33 & 85.18 & 52.47 & 85.61 & 86.47 & 86.21 & 200  \\
        FedKD & 97.74 & 92.58 & 83.22 & 54.16 & 86.32 & 87.36 & 88.22 & 200 \\
        FedGen  & 96.35 & 91.42 & 82.96 & 54.88 & 85.48 & 86.44 & 87.64 & 200  \\ 

        FedBN & 98.56 & 93.10 & 86.37 & 53.35 & 87.54 & 88.63 & 89.48 & 200  \\ 
        FRAug  & 98.17 & 92.58 & 84.51 & 52.67 & 93.64 & 94.51 & 95.16 & 200 \\ 
        FedProto & 98.42 & 93.17 & \textbf{88.15} & 54.42 & 93.16 & 94.16 & 94.23 & 200 \\ 
        FPFL & \textbf{98.68} & \textbf{93.94} & 87.63 & \textbf{60.28} & \textbf{94.75} &\textbf{95.84}& \textbf{96.17}& 200 \\ 
        
    \bottomrule
    \end{tabular}
    }
\label{table_1}
\end{table*}





\begin{figure*}[htbp]
    \centering
    \includegraphics[width=1\textwidth]{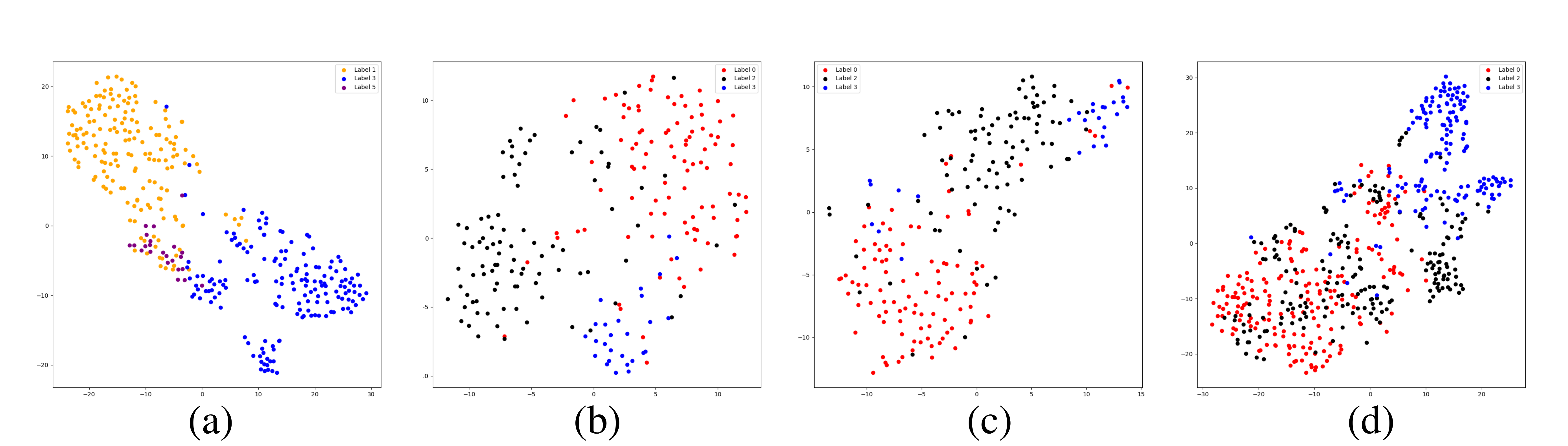} 
    \caption{t-SNE visualization of the prototype generated by the PFPL method. We consider clients from four different domains in the PACS dataset, corresponding to the $(a),(b),(c)$ and $(d)$ in the picture, and the number of classes for each client is uniformly set to $n = 3$.}
    \label{fig_3}
\end{figure*}


\section{Experiments}
\label{sec:5}
\textbf{Datasets:} We evaluate our method on three classification tasks:
\begin{itemize}
    \item[$\bullet$] Digits\cite{roy2018effects} includes four domains: MNIST (M), USPS (U), SVHN (SV), and SYN (SY) with 10 categories (digit numbers from 0 to 9).
    \item[$\bullet$] PACS\cite{li2017deeper} data set is a domain adaptive image dataset, including 4 domains: photos, art paintings, cartoons, and sketches. Each field contains 7 categories.
\end{itemize}

 \textbf{Local models:}
 For these three classification tasks, we use the classical ResNet18 \cite{he2016deep} model as our base model, and all methods use the same network architecture to make fair comparisons across different tasks.

\textbf{Baselines of FL:} We investigate the performance of our method PFPL under mixed heterogeneous conditions and compare it with baselines, including FedAvg, Local. In addition, some FL methods in single-domain scenarios are also included. Feature distribution skewed: FedBN \cite{li2021fedbn}, FRaug\cite{Chen_2023_ICCV}. Label distribution skewed: FedProto\cite{dai2023tackling}, Ditto\cite{li2021ditto}, APFL\cite{deng2020adaptive}, FedRod\cite{li2023adversarial}, FedKD\cite{wu2022communication}.  

\textbf{Mixed heterogeneous setting:} This paper considers a heterogeneous scenario where the label distribution is skewed and the feature distribution is skewed. We borrow the concept of $n$-way, $k$-short from less sample learning, which $n$ controls the number of classes on the client side and $k$ controls the number of training instances per class. To simulate the label distribution skewed, we stochastic change the values of $n$ and $k$ for each client side. For the feature distribution skewed, we stochastic assign instance data from different domains to the client side. The final client-side data only has data for individual category labels and is sourced from different domains, albeit with overlap.

\textbf{ Implementation Details:} We implement the comparison of PFPL and general baseline methods in PyTorch. We use 20 client sides for all data sets. For Digits and PAPC Dataset, the average number of categories $ n$ for local clients is set to 3, 4, 5, and for Office-31 \cite{saenko2010adapting} Dataset, the average number of categories $n$ for local clients is set to 10, 15, 20. And the number of each class in each client side is initially set to 100\%. To make a fair comparison, we follow the same settings. For all methods, we use an SGD optimizer with a learning rate of $ lr = 0.01$. The corresponding weight decay is $ e^{-5} $ and the momentum is 0.9. The training batch size is 4, and we communicate epoch for $ E= 100$  and the local update wheel $ T= 1$.

\textbf{PFPL under varying $\alpha$:} As shown in Equation \ref{eq:1}, in the server level personalized prototype aggregation stage, a controls the weight of the local prototype, as shown in Figure ~\ref{fig_3}, in the range of 0-1, the optimal values of three different data sets a are 0.3, 0.5, and 0.6.

\textbf{The model performance under the number of classes n of different clients:}
Table ~\ref{table_1} reports the average test accuracy for all clients. It can be seen that PFPL has the highest accuracy in most cases among FL under different n controls.
\textbf{Scalability of PFPL on varying number of samples:} 

\begin{figure}
\centering 
\includegraphics[width=1\textwidth]{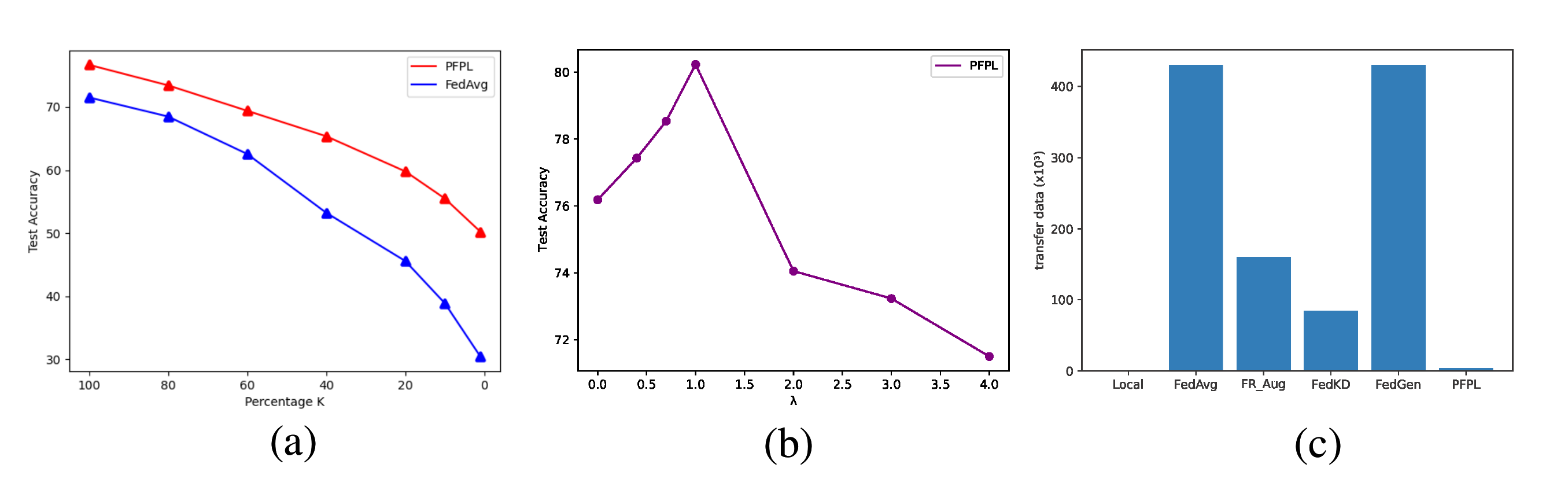} 
    \caption{(a) Average test accuracy of PFPL and FedAvg on PACS with varying numbers of samples in each class. (b) Model accuracy corresponding to different $\lambda$. (c) Comparison of the number of parameters transferred in each round of global iteration} 
    \label{fig_4}
\end{figure} 

Figure \ref{fig_4}a shows that PFPL can scale to scenarios with fewer samples available on clients. The test accuracy consistently decreases when there are fewer samples for training, but PFPL drops more slowly than FedAvg as a result of its adaptability and scalability on various data sizes.

\textbf{PFPL under varying $\lambda$:} Figure \ref{fig_4}b  shows the change in performance at different values of $\lambda$ in Equation ~\ref{eq:5}. We specify the initial range of $\lambda$ at $[0,4]$ and extract a set of values from it. We record the average test accuracy of the PAPC data set, $K = 100\%$, $n = 4$, and the distance loss of the prototype. In this case, as $\lambda$ increases, the original distance loss (regularizer) decreases, while the average test accuracy decreases sharply after $\lambda= 1$, and finally we take the optimal value of $\lambda$ as 1.


\textbf{PFPL communication efficiency comparison:} Figure~\ref{fig_4}c depicts the number of parameters to be transmitted by each client in each communication round. In comparison with the classical approach, our method transmits the minimal number of parameters in each round, thereby effectively minimizing the volume of communication throughout the entire communication round, lowering the communication cost and enhancing the transmission efficiency.

\begin{figure}[htbp]
    \centering
    \subfloat[]{\includegraphics[width=0.48\textwidth]{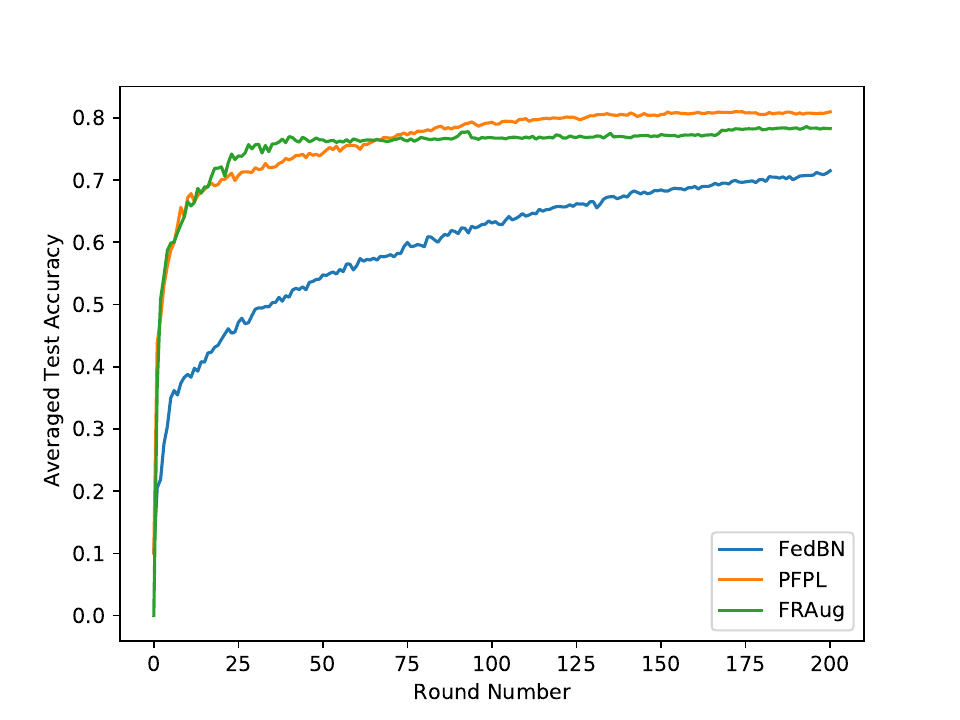}}\label{fig_5a}
    \hfill
    \subfloat[]{\includegraphics[width=0.48\textwidth]{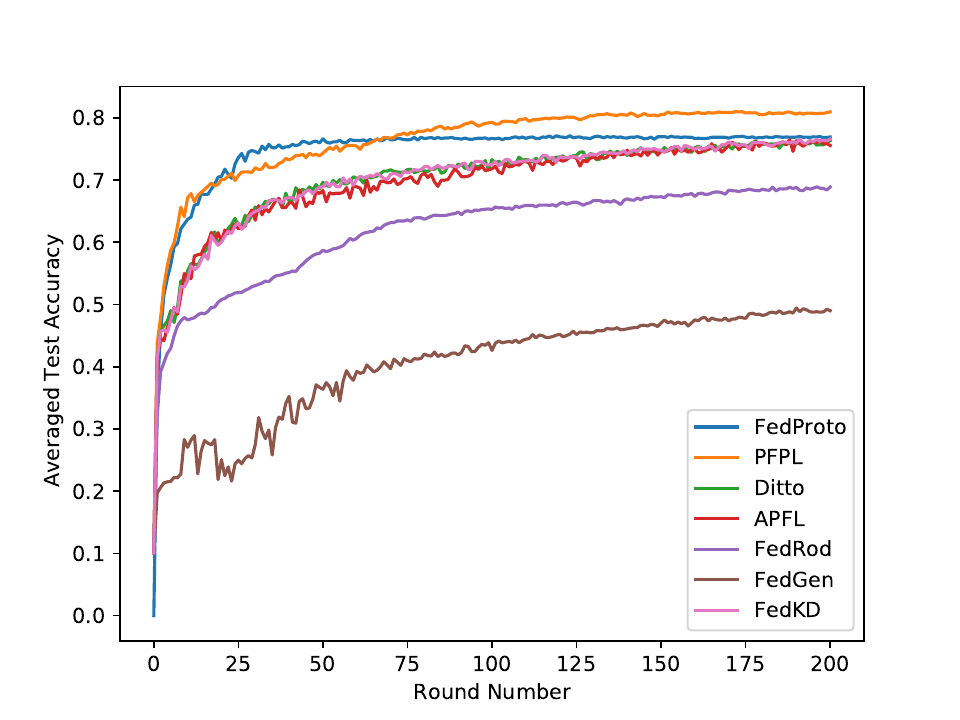}}\label{fig_5d}
    \caption{Precision comparison between PFPL and a single heterogeneous personalized federated learning method in mixed heterogeneous scenarios. (a) indicates the skewed feature distribution scenario, and (b) indicates the skewed label distribution scenario.}
    \label{fig_5}
\end{figure}

\textbf{Performance of PFPL compared to the single-domain FL method:} As shown in Figure~\ref{fig_5} , PFPL achieves higher accuracy than the FL method in single heterogeneous scenarios with skewed label distributions and skewed feature distributions. We suspect that this is due to the fact that previous FL methods focused on solving the heterogeneous federation problem in single scenarios and ignored other heterogeneous scenarios, resulting in lower performance in mixed heterogeneous scenarios.

\section{Conclusion}
\label{sec:6}
In this paper, we explore the personalized federated learning PFPL for handling mixed heterogeneous scenarios. Our work introduces prototype as a communication standard. We use prototype (prototype-like representation) to learn knowledge of different domains and stable convergence goals by generating specific personalized prototypes for different client sides and introducing personalized prototype consistency during the local update phase. Ultimately, our method achieves higher accuracy than single-scenario heterogeneous federated learning methods.

\textit{Feature.} While this study validates PFPL’s~\cite{xing2024personalized} efficacy in mixed heterogeneous classification tasks via personalized prototype-based cross-domain alignment and privacy-preserving communication , its framework holds promising extensions to more computer vision fields. For adverse weather restoration~\cite{fang2025color,zhao2025learning} (e.g., rain, snow, fog), PFPL can address heterogeneous adverse weather data across devices (e.g., surveillance, vehicle-mounted cameras) via client-tailored prototypes, and integrating adversarial training can enhance prototype robustness against extreme weather variations—all while protecting privacy of scene data . In object tracking~\cite{hu2023transformer,hu2025adaptive,hu2025exploiting,hu2024toward,zeng2025explicit,shi2025mamba}, it may fuse distributed heterogeneous features (e.g., RGB, thermal) through prototype aggregation, with adversarial training reducing feature shift impacts to boost tracking stability. For Grounding tasks~\cite{shi2025swimvg}, PFPL’s instance-prototype alignment can unify heterogeneous annotations/features across clients, and adversarial training can refine prototype discrimination, ensuring accurate results without centralized data collection. Future work will adapt PFPL to these tasks and optimize adversarial training~\cite{wang2025fast,wang2024effective} integration for task-specific needs.

\bibliographystyle{splncs04}
\bibliography{ref}





\end{document}